\documentclass[twoside,11pt]{article}

\usepackage{automl2020}

\usepackage{subcaption}
\usepackage{float}

\usepackage[dvipsnames]{xcolor}

\usepackage{authblk}
\author[a]{Roberto L. Castro}
\author[a]{Diego Andrade}
\author[a]{Basilio Fraguela}
\affil[a]{Universidade da Coru\~na, CITIC, Computer Architecture Group. A Coru\~na, Spain}

\firstpageno{1}

\begin{document}

\title{Reusing Trained Layers of Convolutional Neural Networks to Shorten Hyperparameters Tuning Time}



\maketitle

\begin{abstract}
Hyperparameters tuning is a time-consuming approach, particularly when the architecture of the neural network is decided as part of this process. For instance, in convolutional neural networks (CNNs), the selection of the number and the characteristics of the hidden (convolutional) layers may be decided. This implies that the search process involves the training of all these candidate network architectures.

This paper describes a proposal to reuse the weights of hidden (convolutional) layers among different trainings to shorten this process. The rationale is that if a set of convolutional layers have been trained to solve a given problem, the weights calculated in this training may be useful when a new convolutional layer is added to the network architecture. 

This idea has been tested using the CIFAR-10 dataset, testing different CNNs architectures with up to 3 convolutional layers and up to 3 fully connected layers. The experiments compare the training time and the validation loss when reusing and not reusing convolutional layers. They confirm that this strategy reduces the training time while it even increases the accuracy of the resulting neural network. This finding opens up the future possibility of integrating this strategy in existing AutoML methods with the purpose of reducing the total search time.

\end{abstract}

\section{Introduction}

One of the main problems of Automated Machine Learning (AutoML) is the huge amount of time and computational resources that may be consumed by a search process. This is even worse when the search process involves some kind of network architecture selection because the combinatorial explosion skyrockets the number of different alternatives to be tested.

Convolutional Neural Networks (CNNs) have been a traditional target for AutoML techniques, and although the last years have given birth to more exotic variants of the CNN architecture, like inception or yolo, the basic design of this kind of networks is composed of a group of convolutional layers and another group of fully connected ones. This way, when AutoML techniques are applied to CNNs, tuning the number and the characteristics of the convolutional layers is key to increasing the accuracy of the resulting network. This involves search processes where network architectures with different numbers of both convolutional and fully connected layers are tested. This opens the possibility to reuse portions of the network that have been already trained to solve the problem when larger and more complex architectures are trained. We apply this general idea to the reuse of convolutional layers, as it is expected to reduce the training time without sacrificing the accuracy. 

This idea has been tested using the CIFAR-10 dataset~(\cite{krizhevsky2009learning}) exploring two different search spaces: a small one with 252 different hyperparameters values combinations, and a large one with 756 ones. 

The rest of this paper is structured as follows. 
Section~\ref{sec:rw} summarizes the related work. Section~\ref{sec:ssp} describes the layer reuse strategy proposed in this paper. Section~\ref{sec:exp} shows the experiment results, and Section~\ref{sec:conc} presents our conclusions.

\vspace*{-2ex}
\section{Related work}\label{sec:rw}

AutoML is a relevant topic in Machine Learning~(\cite{six,seven}). 
The method used to explore the search space has been a frequent subject of study. The exhaustive exploration strategy, called grid search, is the most systematic one, but it is unfeasible when the search space is too large. As a consequence, strategies such as the Random search narrow the search space by exploring just a fraction of it.
Also, there are more elaborated strategies based on models to traverse more intelligently the search space, like the Bayesian optimization~(\cite{eight}) and the Nelder-Mead method~(\cite{nine}). The use of genetic algorithms is also a trend, not only to tune the hyperparameters, but also to select the most adequate network architecture~(\cite{ten}).

Other works attempt to shorten the search time by reducing the evaluation time of each point of the search space. For instance, early termination of each training is applied in Hyperband~(\cite{eleven}), while in~(\cite{twelve}), the authors exploit the inherent parallelism of the search process to make use of a large amount of computational resources.
Multifidelity optimization techniques~(\cite{thirteen}) explore the idea of simplifying the training process by approximating it.
Successive halving~(\cite{fourteen}) is a variant of this where the exploration process is divided into several stages. Each new stage of the process doubles the training data set while it halves the number of candidate configurations.

Another approach is to reduce the time taken by each individual training by adopting strategies that are specific to a certain kind of neural networks. For instance, in~(\cite{fifteen}), they target CNNs that are first trained using low resolution images, and later, the most promising combinations of hyperparameters are trained again with the full resolution images, initializing the weights of the network with those resulting of the first training process. Our work is focused in this later line of research based on reusing weights of previous trainings. However, our approach differs from this last work in two aspects. First, we reuse the weights not just for the initialization, but we also keep them constant during the whole training process, saving this way time in backpropagation. Second, the architecture of the network in which the weights are reused is different from that of the previous network, usually having an additional convolutional layer.

\vspace*{-2ex}
\section{Layer reuse proposal}\label{sec:ssp}

The idea tested in this paper consists in reusing the weights of convolutional layers between different trainings. This idea may be useful in the context in which we have a network with {\tt n} convolutional layers trained to solve a specific problem, and we want to evaluate whether using an additional convolutional layer may improve the accuracy of the predictions. In a wider context, this could shorten the search time of AutoML processes in which a number of alternative network architectures are tested.



\begin{figure}[httb]
\hspace{-1.1cm}
\center
\includegraphics[width = 10cm]{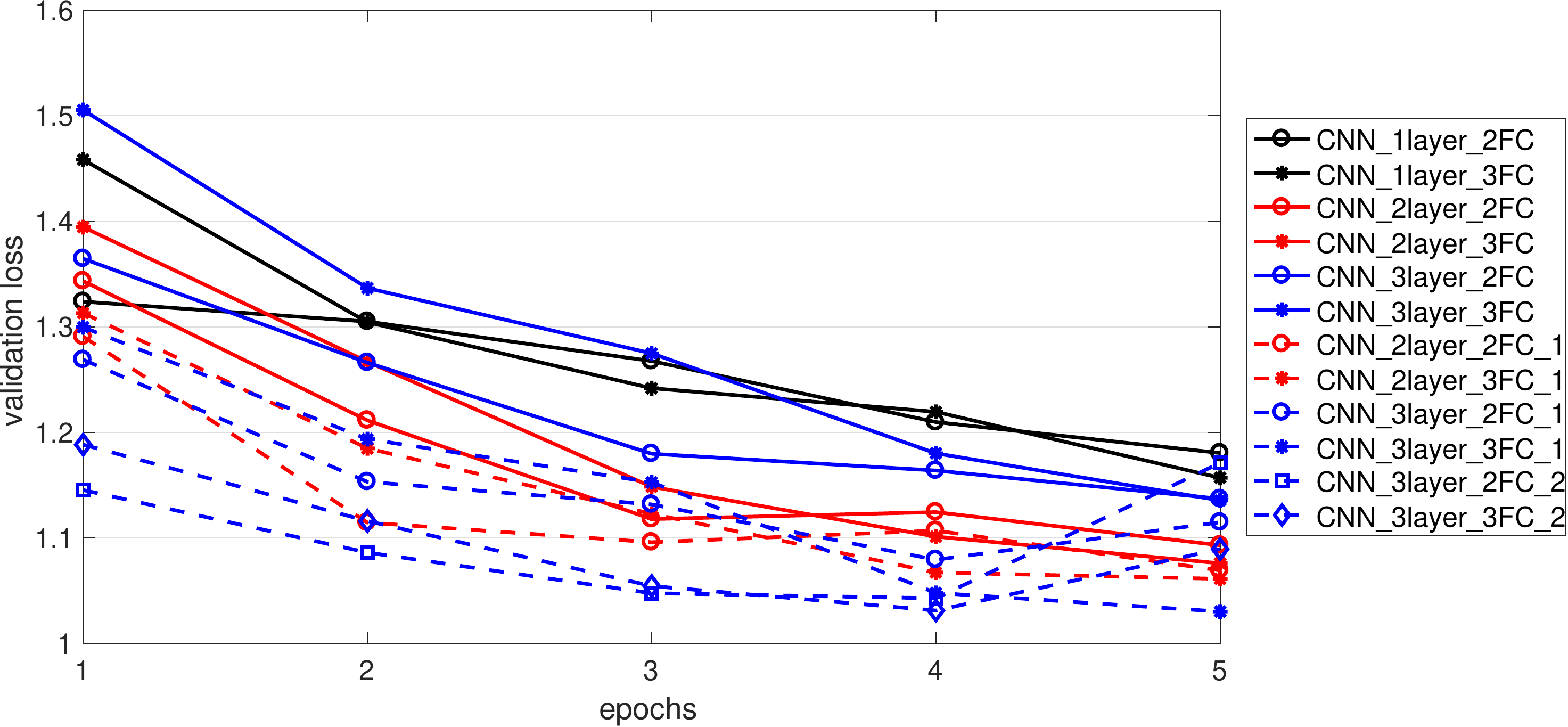}
\
\caption{Effect on accuracy of reusing convolutional layers}
\label{fig:prel}
\end{figure}

First, we evaluate whether the reuse of convolutional layers between different trainings reduces the accuracy of the network.
Figure~\ref{fig:prel} represents the evolution of the validation loss across five training epochs for different network architectures, represented using different colors and patterns. The dataset used for these tests, just as for all the other experiments in this paper, is CIFAR-10. The different networks are named according to the scheme CNN\_\#ncl\_\#nfcl\_\#nrl, where \#ncl is the number of convolutional layers, \#nfcl is the number of fully connected layers and \#nrl is the number of convolutional layers that are reused from a previous training. Notice that the reused layers are the \#nrl first ones. In order to better distinguish between networks with and without reuse, those with reuse are plotted with a dashed pattern. The results clearly show that (1) the reuse of layers not only does not harm but even increases the accuracy of the resulting network, and that (2) a given threshold validation loss can be reached in an earlier epoch with layer reuse. The extensive validation of our hypotheses will be the main topic of Section~\ref{sec:exp}, where a more detailed testing can be found.

We now formalize how the reuse of convolutional layers is performed. In the following, when we talk about reusing a layer, we will be talking about reusing a convolutional (hidden) layer. The same reuse strategy can be applied to reuse fully connected layers, but formalizing and testing this idea is out of the scope of this paper.

The reuse of layers is formalized in the following terms.
There is a trained CNN $R'$, with $\omega$ convolutional layers,
$\zeta$ of which can be reused in the training of another CNN $R$ with $\theta$ convolutional layers, being $\omega< \theta$. The training of the $R$ network will follow these guidelines:

\begin{itemize}
\item The reused layers will be placed consecutively at the beginning of the $R$ network.
\item During the training of $R$, the backpropagation algorithm will only apply to the $\theta-\zeta$ non-reused layers. However, the inference phase will use all the layers.
\item The shapes of the output of the last reused layer and the input of the first non-reused layer must match.
\end{itemize}

Through different experiments, we have also learned that the number of epochs required to train a network with reuse is smaller than those required by the original network. If this is not taken into account, the network with reuse can experience overfitting. 

Also, the network whose layers are reused has to be trained exhaustively, using a significant percentage of the dataset and during a considerable number of epochs. Otherwise, the accuracy of the network with reuse can be affected. 





\vspace{-3ex}

\section{Experiments}\label{sec:exp}




The experiments in this section are designed to validate that the layer reuse strategy reduces the time required by each individual training without harming the accuracy of the resulting network. These experiments use the CIFAR-10 dataset and they explore exhaustively two different hyperparameters search spaces depicted in Table~\ref{tab:ss}, a small one with 252 hyperparameter configurations and a large one with 756. The table shows that these configurations are generated by combining different values of 7 hyperparameters. Five of them affect the architecture of the network: \# of layers, \# of filters, filterSize, \# of hidden layers, and \# units per layer, while the remaining two ones, learning rate and batch size, affect the training process. 

The reuse strategy is compared to a baseline strategy, where all the possible configurations are trained separately with the full dataset. Then, the reuse strategy is applied only to the configurations with more than one layer. In these cases, only the last layer is effectively trained, while the remaining ones are reused from a previous training. The weights of the reused layers are taken from the training of the equivalent configuration in the baseline process.

The tests have been conducted in a cluster with 16 computing nodes, each one equipped with 2 $\times$ {\tt Intel Xeon E5-2660 Sandy Bridge-EP} processors and 64GB of DDR3 RAM. The SMALL test was conducted employing 7 of these nodes, while the LARGE test was conducted using the 16 nodes. In both cases, within each node, 5 physical cores were allocated, each one being assigned 10GB of RAM. The application that run these tests is publicly avaiable \footnote{https://github.com/automl2020/HPAutoML}, and it is built on top of the Ray Python library~(\cite{rayproject_2020}) and Pytorch. The reuse mechanism is implemented in Pytorch through a workaround that avoids backpropagation in the reused layers. The reason for this per-node allocation policy is that Ray requires 10GB of RAM memory per used core, and our system only has 64GB of RAM per node. With these requirements, we could have allocated a maximum of 6 cores per node, but we decided to use just 5 to avoid overloading each node.


\begin{table}[h]
\begin{subtable}{.5\textwidth}
		\centering
		\	\begin{tabular}{lll}
			Hyperparameter & Values \\ \hline
			\# of layers& \{$1, 2, 3$\} \\ 
			learning rate & \{$1e-2, 1e-3$\}  \\
			\# of filters& $18$  \\
			filterSize& $3x3$  \\
			\# of hidden layers & \{$1, 2, 3$\}\\		
			\# units per layer & \{$50, 250, 500$\} \\
			batch size & \{$10, 20, 30, 40$\} \\
		\end{tabular}
		\caption{SMALL: with 252 configurations}
		\label{tab:comb1}
\end{subtable}
\begin{subtable}{.5\textwidth}
		\centering
		\begin{tabular}{ll}
			Hyperparameter & Values \\ \hline
			\# of layers& \{$1, 2, 3$\} \\ 
			learning rate & \{$1e-2, 1e-3, 1e-4$\}  \\
			\# of filters& \{$18, 32, 48, 64$\}  \\
			filterSize& $3x3$  \\
			\# of hidden layers & \{$1, 2, 3$\}\\
			\# units per layer & \{$50, 250, 500$\} \\
			batch size & \{$20, 30, 40$\} \\
		\end{tabular}
		\caption{LARGE: with 756 configurations}
		\label{tab:comb2}
\end{subtable}
	\caption{Configuration of the traversed search spaces}
	\label{tab:ss}
\end{table}

\begin{figure}[H]
	\begin{subfigure}{.5\textwidth}
		\centering
		\includegraphics[width=.9\linewidth]{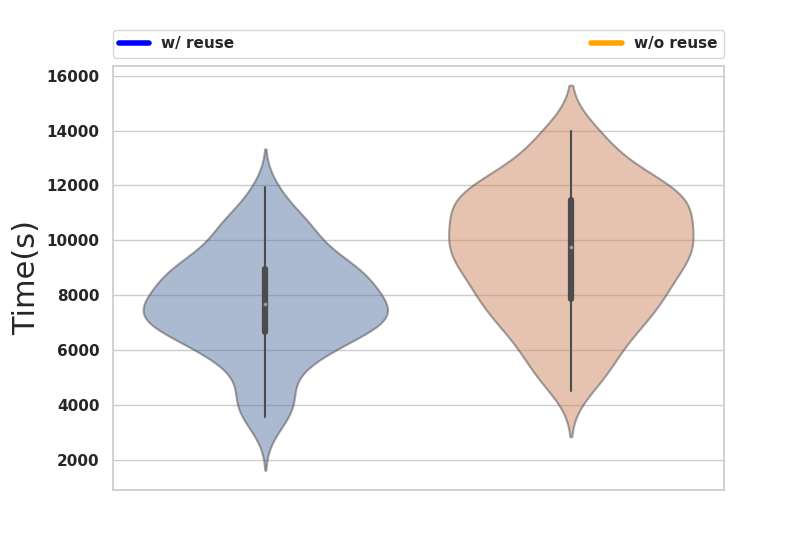}
		\caption{Statistical distribution of the training time}
		\label{fig:smalltime}
\end{subfigure}
\begin{subfigure}{.5\textwidth}
		\centering
		\includegraphics[width=.9\linewidth]{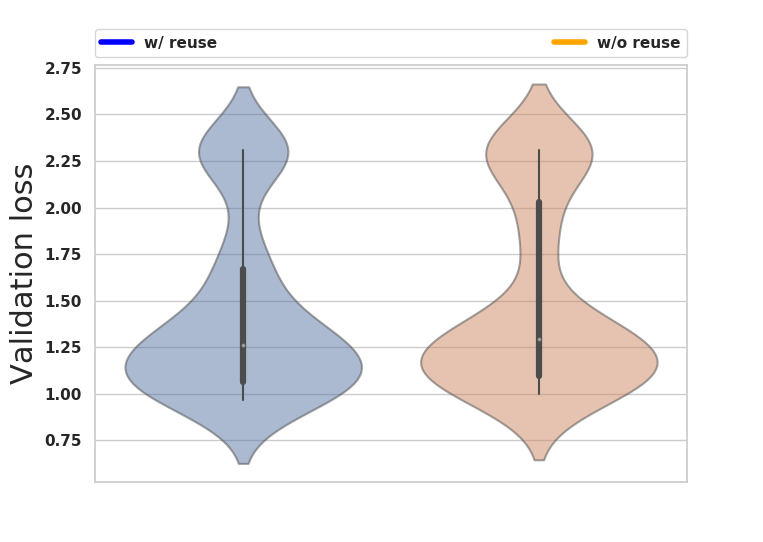}
		\caption{Statistical distribution of the validation error}
		\label{fig:smallvl}
	\end{subfigure}
		\caption{SMALL case}
		\label{fig:small}
\end{figure}

\begin{figure}[H]
	\begin{subfigure}{.5\textwidth}
		\centering
		\includegraphics[width=.9\linewidth]{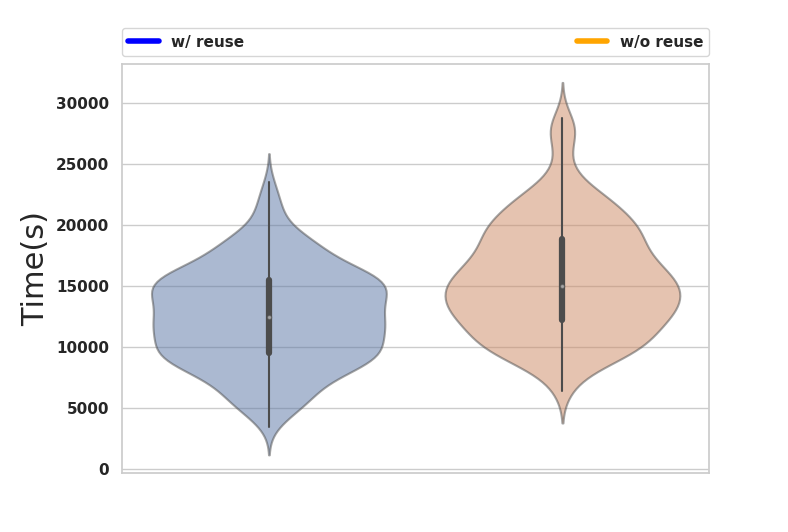}
		\caption{Statistical distribution of the training time}
		\label{fig:largetime}
\end{subfigure}
\begin{subfigure}{.5\textwidth}
		\centering
		\includegraphics[width=.9\linewidth]{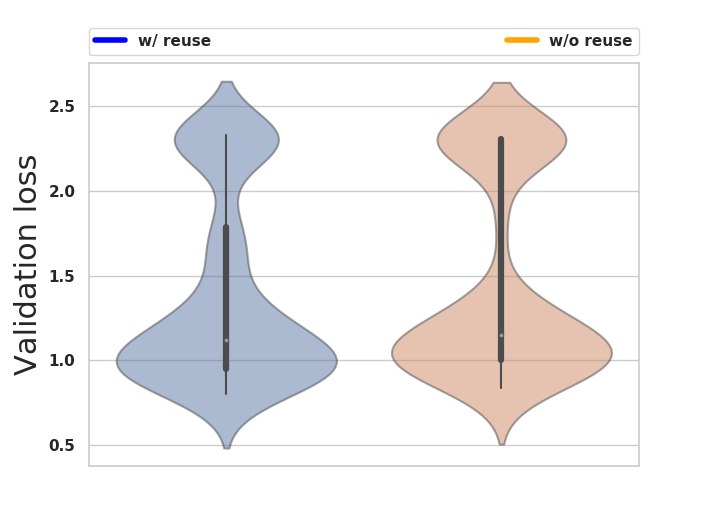}
		\caption{Statistical distribution of the validation error}
		\label{fig:largevl}
	\end{subfigure}
		\caption{LARGE case}
		\label{fig:large}
\end{figure}

Figures~\ref{fig:smalltime} and~\ref{fig:smallvl} represent the statistical distribution of the training time and the validation loss, respectively,  when reusing and when not reusing layers, across 168 of the 252 configurations of the SMALL test. 
This figure excludes the 1-layer configurations because it is not possible to reuse layers in these configurations. The results clearly show that although both distributions have a similar shape, the reuse strategy considerably reduces the training time while it reduces the error of the resulting network.
 The total training time for all the cases with more than one layer without reusing is $14000$ seconds and with reuse $11000$ seconds. This implies a  total saving in the search time of $50$ minutes. Regarding the validation loss, the value of the best configuration using reuse is $0.9651$, while without reuse it is $1.0016$.
 
 An analogous experiment was done with the LARGE case and the results are reflected in Figures~\ref{fig:largetime} and~\ref{fig:largevl}, respectively. Once again, $504$ of $756$ configurations are represented as one-layer configurations are once again excluded. The total training time for all the cases with more than one layer without reusing is $30000$ seconds and with reuse it is $23000$ seconds. This means a reduction of $1$ hour and $56$ minutes. The validation loss of the best configuration using reuse is $0.85$, while without reuse it is $0.90$.

Following the nomenclature:
\begin{itemize}
	\item \textit{\#convlLayers, learningRate, \#filters, filterSize, \#hiddenLayers, hiddenLayersUnits, batchSize}
\end{itemize}

The best model configuration in the SMALL tests without reuse was: 3, 0.001, 18, 3, 2, [500, 250], 40
while the best model configuration with reuse was: 3, 0.001,18, 3 ,1, [250], 40. In the case of the LARGE tests the best configuration without reuse was: 3, 0.001, 48, 3, 2, [500,250], 40; while the best one with reuse was: 3, 0.001, 48, 3, 2, [500,50], 30.

This shows that although the best configurations both with and without reuse are very similar, they are not exactly the same. This makes sense, as the networks, although structurally similar, are trained using a different initial situation and following a different procedure.

\begin{table}[H]
	\centering
	\begin{tabular}{l|ll|l|}
		Epoch &  w/o reuse &&   w/ reuse   \\ \hline
    	1 & 1.16  &&  0.93   \\
    	2 & 1.03  &&  0.85   \\
    	3 & 0.97  &&  \textcolor{red}{0.85}  \\
    	4 & 0.96  &&  \textcolor{red}{0.86}   \\
    	5 & 0.90  &&  \textcolor{red}{0.87}   \\\hline
	\end{tabular}	
	\caption{Validation loss evolution during 5 \textit{epochs}}
	\label{tab:others2}
\end{table}

Finally, Table~\ref{tab:others2} shows the evolution of the validation loss during the 5 training epochs of the best configurations with and without reuse in the LARGE tests. As we can see, the training with reuse achieves the minimum validation loss earlier. This allows detecting stabilizations or increases in the validation loss, which could mean that the net is being overfitted. As we have explained, reusing layers tends exhibits this behavior (approximately in around a ~50\% of the configurations), so the training can be stopped earlier, thus reducing the number of epochs required.


\vspace*{-1ex}

\section{Conclusions}\label{sec:conc}

We have shown that reusing convolutional layers in CNNs between different trainings reduces training time without affecting the accuracy of the resulting network. This opens the possibility of incorporating the layer reuse strategy in large AutoML search processes where the architecture of the neural network is also decided, as a mean to save time without sacrificing accuracy.

As future work, it is crucial to design search strategies specially designed to take advantage of this reuse mechanism.  Early experiments, not included in this paper, have shown that the existing strategies cannot directly benefit from it, but it is possible to design reuse-aware search strategies that can help us reduce the search time. 



\acks{}
El CITIC, como Centro de Investigaci\'on del Sistema universitario de Galicia es financiado por la Conseller\'ia de Educaci\'on, Universidade e Formaci\'on Profesional de la Xunta de Galicia a trav\'es del Fondo Europeo de Desarrollo Regional (FEDER) con un 80\%, Programa operativo FEDER Galicia 2014-2020 y el 20\% restante de la Secretar\'ia Xeral de Universidades (Ref ED431G 2019/01).

\vskip 0.2in
\bibliography{main}

\begin{thebibliography}{12}
\providecommand{\natexlab}[1]{#1}
\providecommand{\url}[1]{\texttt{#1}}
\expandafter\ifx\csname urlstyle\endcsname\relax
  \providecommand{\doi}[1]{doi: #1}\else
  \providecommand{\doi}{doi: \begingroup \urlstyle{rm}\Url}\fi

\bibitem[Hinz et~al.(2018)Hinz, Navarro-Guerrero, Magg, and Wermter]{fifteen}
Tobias Hinz, Nicol{\'a}s Navarro-Guerrero, Sven Magg, and Stefan Wermter.
\newblock Speeding up the hyperparameter optimization of deep convolutional
  neural networks.
\newblock \emph{International Journal of Computational Intelligence and
  Applications}, 17\penalty0 (02):\penalty0 1850008, 2018.

\bibitem[Hutter et~al.(2019)Hutter, Kotthoff, and Vanschoren]{six}
Frank Hutter, Lars Kotthoff, and J.~Vanschoren, editors.
\newblock \emph{Automatic machine learning: methods, systems, challenges}.
\newblock Challenges in Machine Learning. Springer, Germany, 2019.
\newblock ISBN 978-3-030-05317-8.
\newblock \doi{10.1007/978-3-030-05318-5}.

\bibitem[Johnson and McCourt(2018)]{twelve}
Alexandra Johnson and Michael McCourt.
\newblock Orchestrate: Infrastructure for enabling parallelism during
  hyperparameter optimization.
\newblock \emph{arXiv preprint arXiv:1812.07751}, 2018.

\bibitem[Kandasamy et~al.(2016)Kandasamy, Dasarathy, Oliva, Schneider, and
  P{\'o}czos]{thirteen}
Kirthevasan Kandasamy, Gautam Dasarathy, Junier~B Oliva, Jeff Schneider, and
  Barnab{\'a}s P{\'o}czos.
\newblock Gaussian process bandit optimisation with multi-fidelity evaluations.
\newblock In \emph{Advances in Neural Information Processing Systems}, pages
  992--1000, 2016.

\bibitem[Klein et~al.(2016)Klein, Falkner, Bartels, Hennig, and Hutter]{eight}
Aaron Klein, Stefan Falkner, Simon Bartels, Philipp Hennig, and Frank Hutter.
\newblock Fast bayesian optimization of machine learning hyperparameters on
  large datasets.
\newblock \emph{arXiv preprint arXiv:1605.07079}, 2016.

\bibitem[Krizhevsky et~al.(2009)Krizhevsky, Hinton,
  et~al.]{krizhevsky2009learning}
Alex Krizhevsky, Geoffrey Hinton, et~al.
\newblock Learning multiple layers of features from tiny images.
\newblock 2009.

\bibitem[Kumar et~al.(2018)Kumar, Dahl, Vasudevan, and Norouzi]{fourteen}
Manoj Kumar, George~E Dahl, Vijay Vasudevan, and Mohammad Norouzi.
\newblock Parallel architecture and hyperparameter search via successive
  halving and classification.
\newblock \emph{arXiv preprint arXiv:1805.10255}, 2018.

\bibitem[Li et~al.(2017)Li, Jamieson, DeSalvo, Rostamizadeh, and
  Talwalkar]{eleven}
Lisha Li, Kevin Jamieson, Giulia DeSalvo, Afshin Rostamizadeh, and Ameet
  Talwalkar.
\newblock Hyperband: A novel bandit-based approach to hyperparameter
  optimization.
\newblock \emph{The Journal of Machine Learning Research}, 18\penalty0
  (1):\penalty0 6765--6816, 2017.

\bibitem[Nazir et~al.(2020)Nazir, Patel, and Patel]{seven}
Sajid Nazir, Shushma Patel, and Dilip Patel.
\newblock Assessing hyper parameter optimization and speedup for convolutional
  neural networks.
\newblock \emph{International Journal of Artificial Intelligence and Machine
  Learning (IJAIML)}, 2020.

\bibitem[Ozaki et~al.(2017)Ozaki, Yano, and Onishi]{nine}
Yoshihiko Ozaki, Masaki Yano, and Masaki Onishi.
\newblock Effective hyperparameter optimization using nelder-mead method in
  deep learning.
\newblock \emph{IPSJ Transactions on Computer Vision and Applications},
  9\penalty0 (1):\penalty0 20, 2017.

\bibitem[repository(2020)]{rayproject_2020}
Ray~Project repository.
\newblock ray-project/ray, May 2020.
\newblock URL \url{https://github.com/ray-project/ray}.

\bibitem[Young et~al.(2015)Young, Rose, Karnowski, Lim, and Patton]{ten}
Steven~R Young, Derek~C Rose, Thomas~P Karnowski, Seung-Hwan Lim, and Robert~M
  Patton.
\newblock Optimizing deep learning hyper-parameters through an evolutionary
  algorithm.
\newblock In \emph{Proceedings of the Workshop on Machine Learning in
  High-Performance Computing Environments}, pages 1--5, 2015.

\end{thebibliography}






\end{document}